# Alexa, Predict My Flight Delay

Sia Gholami [a*], Saba Khashe [b]

[a,b] *The Institute of Electrical and Electronics Engineers, Member IEEE*
[a]*Email: gholami@ieee.org*

**Abstract**

Airlines are critical today for carrying people and commodities on time. Any delay in the schedule of these planes can potentially disrupt the business and trade of thousands of employees at any given time. Therefore, precise flight delay prediction is beneficial for the aviation industry and passenger travel. Recent research has focused on using artificial intelligence algorithms to predict the possibility of flight delays. Earlier prediction algorithms were designed for a specific air route or airfield. Many present flight delay prediction algorithms rely on tiny samples and are challenging to understand, allowing almost no room for machine learning implementation. This research study develops a flight delay prediction system by analyzing data from domestic flights inside the United States of America. The proposed models learn about the factors that cause flight delays and cancellations and the link between departure and arrival delays.

**1. Introduction**

Air transportation has increased the number of flights, resulting in increasingly severe delayed flights. Flight delays interrupt passengers' travel arrangements and result in significant financial losses for airline companies. Consider the United States: the overall cost of all commercial aviation disruptions in 2008 was $33.5 billion [1]. According to air carrier delay statistics, only 79 percent of flights arrived on time in 2019, resulting in tens of billions of dollars in losses, including costs to airline passengers, demand loss, and other indirect expenses [2]. Flight delay prediction is critical for reducing flight delays. Creating a reliable and accurate flight delay predictive model can provide decision-making with a clear path to achieving effective scheduling decisions. As a result, airline companies and research scientists are focusing on decreasing flight delays.

With the rapid growth of air transportation, flight delays have become a major concern for both airlines and customers. Passengers also waste time, and they often lose faith in airlines. National flight delays cost the United States 31.2 billion dollars, according to Nextor, with 8.2 billion dollars explicitly affecting airlines, 16.2 billion dollars affecting passengers, and 4.0 billion dollars in deferred GDP. It has also been revealed that 33 percent of overall concerns raised from customers are related to flight delays. Airlines are suffering significant financial losses, and their reputation is also suffering. Consequently, it's crucial to properly analyze and anticipate aircraft delays [3]. Therefore, an automatic system is required to assist airline companies and passengers in timely prediction for their flight bookings to save precious time.

---

[*] Corresponding author.

The advancement of artificial intelligence (AI) can be helpful for airline companies in optimizing the performance of existing airports, as the unutilized airspace and airport volume can be designated for a new flight. The upsurge and growth of AI technologies, particularly machine learning (ML) and deep learning (DL) algorithms allow us to more efficiently determine important information in massive data. It helps us in the comprehension and prediction of many more complicated patterns [4] and is widely used in various disciplines such as business decisions, computer vision, natural language processing, and automatic driving. Furthermore, DL is now used to predict ground traffic flow [5]. It is worth the effort to assess the relevance and efficiency of a DL approach enabling flight delay prediction, which is one of the air travel data analytics applications.

We used supervised machine learning techniques to investigate individual flight arrival delays in this study. Machine learning is repeatedly tried for several motives; for beginners, the amount of actual flight and the seasonal database has become too enormous to be analytically examined [6]. Furthermore, it is challenging to test all assumptions due to the very complicated and nonlinear interactions between causal aspects, delay, and correlations across elements. With large volumes of information, ML can quickly develop frameworks and discover and exhibit invisible patterns from the data. To summarize, ML is an intelligent approach for addressing issues in simulation and analytical calculation with a vast volume of information.

Many DL and ML architectures, including stacked autoencoders, CNNs, and RNNs, are used for prediction worldwide. In this research study, various machine learning and deep learning models are created using a dataset containing flight data from prior years. Several analytical methods are utilized in these models to decide which machine learning model best matches the model on delay and cancellation prediction. The purpose is to establish a delay prediction model based on a machine learning algorithm to predict the departure delay at an airport. Our contributions are as follows:

- Collection, preprocessing, and feature engineering of the Bureau of Transportation Statistics (US Department of Transportation) flights dataset.

- We developed three different models to investigate and quantify a more comprehensive range of factors that may potentially influence flight delays.

- Implementation of a conversational voice-based agent to interact with users.

The rest of this article is organized as follows: A summary of the significant results on flight delay prediction is given in Section 2. Data processing and related characteristics are introduced in Section 3. Section 4 develops a prediction model for flight delays, while Section 5 examines overall features' contributions to the prediction findings. Section 6 sums up the research and makes recommendations for further work.

## 2. Related Work

Artificial intelligence, machine learning, and deep learning has been a major area of research since the early days of computer science [7], [8], [9], [10], [11], [12], [13], [14]. Balamurugan et al. [15] proposed several learning methods for forecasting airline cancellations and delays. A random forest regressor with the most significant number of leaf nodes in each tree structure set to 350 may best predict flight delays and achieve a validation MSE of 0.5. A random forest classifier with a mean AUC score of 0.639 is the better model for predicting a particular flight. Gui et al. [16] proposed the Random Forest regresses approach, which was chosen as the best for Arrival Delay, with the minor overall average error 2261.8 and the absolute value of standard error 24.1. The with lowest mean square (RMS Errors 3019.3 and mean absolute error 30.8, the Random Forest regresses were the best model in Arrival Delay. Random Forest-based and LSTM-based architectures were used by Zang et al. [17] to anticipate flight delays. The experimental findings suggest that the Random Forest-based technique performs well for binary classification tasks, but there is still potential for improvement for multi-category classification tasks. The training accuracy of the long short-term memory architecture is considerably higher, showing that the LSTM unit is an appropriate framework for dealing with data streams.

Chen Tan [18] included flight operation records and meteorological data. They developed a delayed flight prediction built on the Catboost technique to estimate the departure delay at the terminal. The model's reliability is demonstrated with NLIA data in 2017 and 2018. Moreover, the model's prediction accuracy is 0.77, according to the results. Jiang et al. [19] presented a thorough aviation data analysis on aircraft delays. QCLCD and AOTP data are combined to create a new database that includes flight details and weather conditions. The dataset is examined further, revealing a helpful connection regarding flight delay. Then, prediction models based on machine learning approaches are created and compared, with the Multilayer Perceptron network achieving the highest accuracy of 89.07 percent. Kalyani et al. [20] used machine learning methods such as the decision tree algorithm (XGBoost) and linear regression to anticipate flight delays and estimate delay duration. The researchers used data collection of both aircraft and meteorological information to evaluate the specified inputs and validate them using machine learning classification and regression techniques.

Yiu et al. [21] used multiple machine learning approaches to anticipate aircraft delays and compare their results in the HKfA scenario. According to the findings, the ANNs model is the most successful in predicting flight delays. While Nave Bayes is the least effective, CNN has the lowest F1 score. Nonetheless, the overall accuracy of different methods was at least 80%. They have provided promising results, indicating that these systems may generally anticipate flight delays. Kim et al. [22] deployed a deep learning model that can increase the accuracy of airline delay prediction. An effective delay condition of different days can be received using an LSTM RNN framework for the prediction algorithm. The researchers obtained the most precise statuses for individual flights by applying an algorithm to flight status. It predicts flight delays with cutting-edge accuracy. The next stage uses different deep architectures to predict and analyze flight delays. It may reveal significant trends in flight delay data. Venkatesh et al. [23] demonstrates that neural networks and multilayer perceptron predict delayed flights efficiently. The model's accuracy using neural networks is 93% by adding one dense layer containing three neurons. The suggested method is evaluated on a large dataset of real-world flight data, with an accuracy of 77 percent using deeper networks and 89 percent using neural networks.

**3. Data**

From 2010 to 2020, from the Airline On-time Performance database of the Bureau of Transportation Statistics in the US department of Transportation, we obtained information on domestic aviation operations in the US and weather information. The Bureau of Transportation database includes statistics on the internal direct flights offered by airline companies. It also offers data on origin airports, airplane identification numbers, schedules, and delay durations. The National Ocean and Atmospheric Administrations' dataset comprises meteorological data such as wind, cloud level, transparency, temperature, pressure, rainfall, and so on, which are generally recorded hourly at global stations. Choosing a set of valuable attributes for the system from various attributes is based on feature selection. Null value cleaning was carried out before feature selection. Additionally, it offers details such as aircraft numbers, timetables, origin/destination airlines, and delay durations.

*3.1. Data Preprocessing*

On-time flights depart at the airport within fifteen minutes of the scheduled time. Flights that were delayed or redirected in the training set were considered delayed flights. Linear interpolation with two adjacent known values compensates for missing values in meteorological data. Only the label Arr Del 15 has missing values in the dataset. As a result, cases with missing tags were eliminated from the dataset. Many of the categorical attributes listed are label encoded, with numbers beginning with 0 allocated alphabetically to each category. For features, we used One-Hot Encoding, which requires categorizing different categorical features into categories. The subcategories are different values taken by them, with each type allocated a binary output, 0 if something doesn't adhere to a particular group, and one if it does. This is done for categorical features with fewer unique values or categories to prevent the computational burden. For the past two years, data has been used for training, wherein 1% of data is used for testing the model.

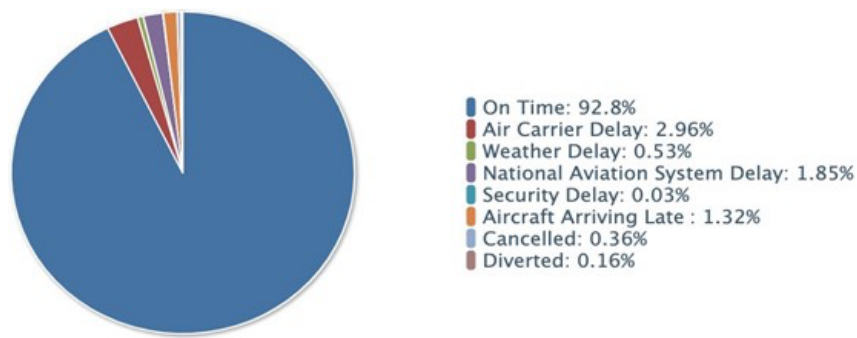

**Figure 1:** Database of the Bureau of Transportation Statistics in the US department of Transportation.

*3.2. Features*

Every characteristic in the flight database is significantly essential, as shown in Table 1. We removed several features from the dataset for a variety of reasons. F1 is a minimal variation that contains flight data for two years. F2 seems to be a relatively repetitious attribute because of the monthly information. Additionally, as the dataset retrieves originating and destinations locations from characteristics F7 and F8, features look repetitive, respectively. However, there may be cases where more than one airport has the corresponding global identifier; thus, F9 was discarded to retain both the adaptability and clarity of our algorithm.

**Table 1:** Features of the Bureau of Transportation Statistics dataset.

| Attribute/Feature Name | ID | Attribute Type |
| --- | --- | --- |
| Year | F1 | Categorical |
| Quarter | F2 | Categorical |
| Month | F3 | Categorical |
| Day of Month | F4 | Categorical |
| Day of Week | F5 | Categorical |
| Flight Num | F6 | Categorical |
| Origin Airport ID | F7 | Categorical |
| Destination Airport ID | F8 | Categorical |
| Destination World Area Code | F9 | Categorical |
| CRS Departure Time | F10 | Continuous |
| CRS Arrival Time | F11 | Continuous |

## 4. Approach

We implemented three different algorithms for flight delay prediction based on ML and DL. The overall structures and implementation are discussed in detail in section 4.1 to 4.5.

*4.1. Model 1 Carrier Origin (Linear Regression)*

It is critical to determine relevant factors that impact flight cancellation or delays and the user information to predict them. After conducting extensive data analysis, we discovered a close relationship between arrival and departure delay. As a result, we may utilize the departure delay to estimate the arrival delay. To solve the issue, we develop a simple linear regression algorithm.

$$Y = \beta_1 X_1 + \beta_2 X_2$$

$Y$ in this case stands for arrival delay, $X_1$ for departure delay, and $X_2$ for path mileage. The analyses in this research used the control variables: schedule and type of aircraft, real departing times, departure airport, travel mileage, and weather at the destination airport. During the training stage, we apply the origin city and airline training models to learn the two parameters $\beta_1$ and $\beta_2$. The path distance for a flight leaving through one airport is already determined at the prediction phase. If we know the flight's departure delay, we may use the model to predict its arrival delay, allowing us to decide whether the flight is delayed or not. We initially look for comparable information in the database based on the departure airport, weather, and the type of aircraft at the departure airport. Finally, utilizing departure delay, we combine the very same information and choose a threshold. We calculate the average departure delay and use it as the plane's departure delay if the volume of information reaches a particular threshold. The whole computational framework is shown in Figure 2.

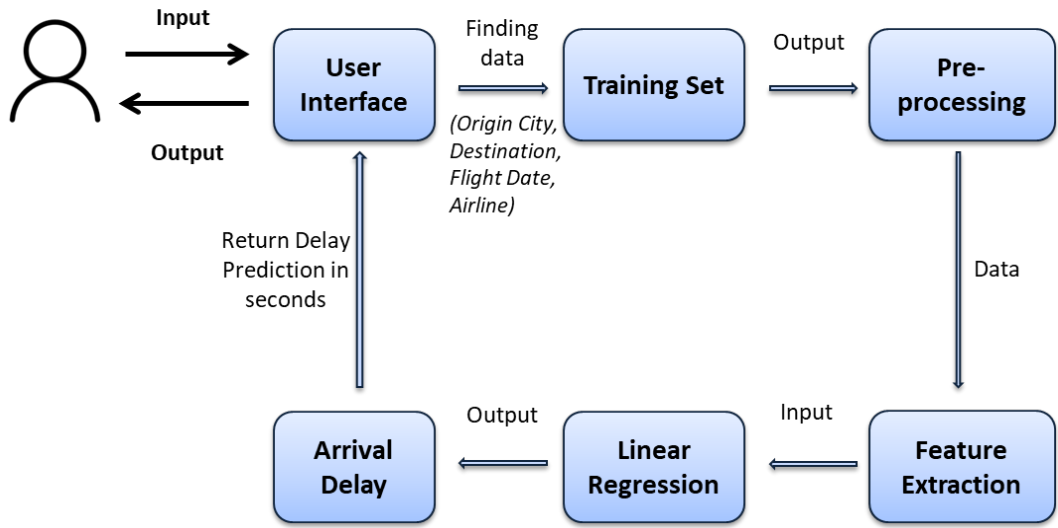

**Figure 2:** Overall Framework of Carrier Origin Model based on linear regression.

*4.2. Model 2 Seasonal (Decision Tree + Linear Regression)*

The model is a simple decision tree plus four different linear regression models. The decision tree will determine the prediction season (e.g., spring) and then use the corresponding linear model for prediction. We trained four other models independently based on their appropriate seasonal data points. The decision tree is one of the simplest and most straightforward predictive modeling tools. Decision trees stratify or partition the predictor space into numerous essential areas to generate a prediction for a given observation. The mean or median is used in the training data of the corresponding region. The decision tree for the aircraft data set is depicted in the figure (3). The most essential or relevant variable/node is at the top. In this scenario, the value in the square boxes represents the forecast of the departure delay estimation.

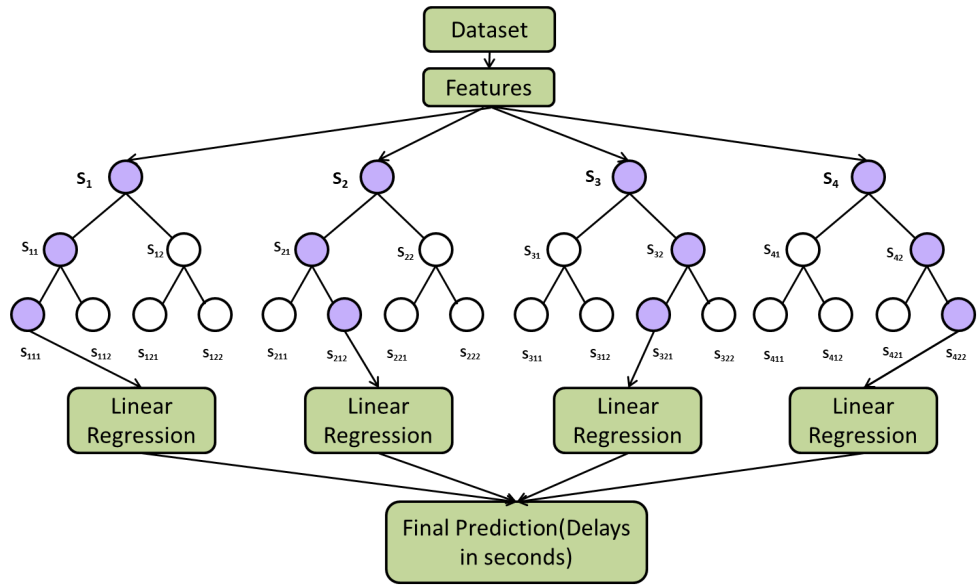

**Figure 3:** Framework of Seasonal Model based on decision tree and linear regression algorithms.

*4.3. Model 3 (Multi-Layer Perceptron)*

The final model was a 5-layer MLP multilayer perceptron trained on all the data. Within the scope of the study, we utilized an artificial neural network (ANN) as a supervised learning model. ANN may be seen as a weighted directed graph, with artificial neurons serving as nodes and weighted directional edges connecting neuron inputs and outputs. In ANN, the designs might vary depending on the connecting patterns, and the input or output layers can be more transitional layers because of their dependability and superior performance. We used the multilayer perceptron (MLP) to estimate traffic arrival delays using ANN. MLP, unlike other statistical approaches, can model extremely nonlinear relationships and has been proven beneficial when confronted with previously unexplored data.

$$Z = f(b + xi.wi)$$

The MLP has been used for various applications, including prediction, function approximation, and pattern categorization. The final model consists of five hidden layers (300,200,100,50 neurons) followed by a linear activation function. Figure 4 shows the overall structure of the MLP model.

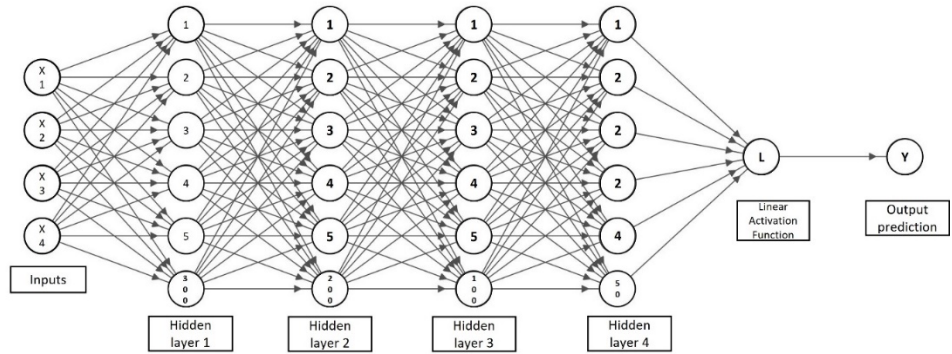

**Figure 2:** Multi-layer perceptron structure trained on Bureau of Transportation Information dataset.

*4.4. User Experience Research*

We conducted exploratory research including end-user interviews and comparative analysis to explore inputs and outputs of the system. Through an iterative process, we discussed possible design ideas in the team and generated design concepts, tested the concepts with potential users, gathered users' feedback, and improved the design based on the test results and users' feedback.

Our research questions are as follows:

- What information is needed to provide predictions for the future flights?
- How to feed the information to the system?
- How should the outputs be presented?
- What do users expect a flight status tool to do?

We interviewed 12 frequent flyers at the airports whose flights were delayed, asking about their experience of flight delays. We asked them to show the methods they used to find information about the status of their flights before coming to the airport, and we asked the following questions:

- What information did they want to have about their flight status?
- What was frustrating using the existing tools?
- What could be better about the existing products?

We have found that users need to have data about their upcoming flights and flight delay predictions. In users' experience, flight delays are highly dependent on airline and origin city; we also found out that voice commands are preferred over filling in forms.

Based on end-user interviews, we defined the user experience of the system:

- We used linear regression and neural network models to make predictions about flight delays based on past

flight data.
- Amazon Alexa was used to communicate with users.
  - The menu options were limited to four to make sure that users would not forget them since voice options were used.
- The software was designed to provide the following options:
  - List existing flights
  - Check a flight from the existing list
  - Add or remove flights from the list
  - Get flight delay information

We conducted usability tests with 15 graduate students at USC Viterbi School of Engineering using some task scenarios. An example of the task scenario: "You've planned a vacation for the New Year. You have purchased your plane ticket, and now you want to have delay information about your flight. Use Flight Stat to find information about your flight."

Participants had to answer the following questions about each flight that they wanted to add, remove or get its delay information:

- Where were they flying from and flying to?
- What airline were they flying with?
- What date and time was their flight?

In addition, if they wanted to get information about one of the listed flights, they had to answer one of the following questions:

- Did they want to check the delay of their next flight?
- Did they want to check a flight's delay with the origin city?
- Did they want to check a flight's delay with date and time?

Based on the information provided, Alexa could add or remove a flight to the list, check a flight, list flights, or provide a flight delay information.

We found the following usability issues:

- When Alexa asked a user to provide both origin and destination at the same time, the following errors happened:
  - Users forgot to provide the information regarding both cities or airports (they just provided the origin or destination).
  - Users did not provide the data in the right order (destination was mentioned first instead of origin).

- When Alexa asked a user to provide both date and time at the same time, users forgot to provide information about the time.

These issues resulted in:

- More lines being exchanged between Alexa and users in cases that users forgot to provide complete answers.
- The wrong answer was provided by Alexa when users did not answer the questions in the right order.

Based on this research, we made the following refinements to the app:

- Removed memory burdens by asking questions one at a time:
  - Where are you flying from?
  - Where are you flying to?
  - What airline?
  - When are you flying?
  - What time is your flight?
- Added greetings and closing comments

### 4.5. Implementation

The user interface is implemented as an Alexa application that will get input such as (as flight source, destination, Origin, and departure time) from the user. It is a cloud-based, voice-activated assistant available in Amazon Echo, Echo Dot, and other products that interact with the user to ask about their query. The Amazon API gateway will trigger the lambda function known as Flight Stat. Lambda function operates as a streaming client, accepting an index of the imported information as a unique event. The lambda function does further computation. Each ML and neural network model are packed in a lambda function that the FlighStat lambda function calls remotely and simultaneously. The three lambda functions contain three different models (Carrier/Origin, Seasonal and Neural Net) trained on the Bureau of Transportation Statistics in US department of Transportation dataset for flight prediction delays. Amazon ML hosts the seasonal and carrier/origin algorithms serverless. All pre-trained algorithms are not included in the Alexa request process, and the S3 bucket contains the weights of pre-trained models used in these lambda functions. These three lambda functions will return a prediction, which will be sent to the user through the Alexa app.

On the other side, the flight data is stored in a DynamoDB table with streams enabled. Amazon DynamoDB offers quick and versatile NoSQL database distribution services and triggers that may be integrated with AWS Lambda to enable similar data to be accessible everywhere. And the payload from DynamoDB will be written to an Amazon Kinesis data stream using a lambda function. The data is stored in Amazon Kinesis until a certain level is crossed. At this point, it will be sent to Kinesis Firehose to be converted from raw format to Elasticsearch-compatible format. Kibana is a log and time-series web analytics visualization and investigation tool. It enables the customers to access Elasticsearch databases for information and then checks the results using basic chart choices or built-in applications. Here the backend user can access the comprehensive analytics using the Kibana dashboard.

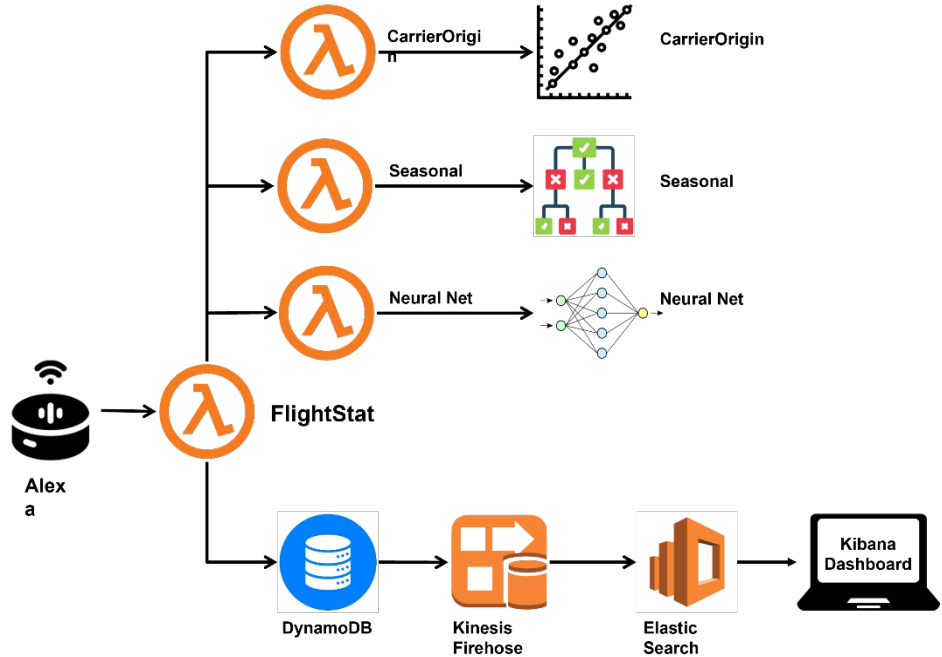

**Figure 3:** Overall Implementation of Carrier Origin, Seasonal and MLP models using Amazon serverless.

## 5. Experiments

We evaluated our models using R-squared technique. The R-squared technique is a statistical measure of how well one set of data can be used to predict another in regression models. It is also called coefficient of determination or coefficient of multiple determination, and it is denoted by the letter R-squared.

$$R^2 = \frac{\sum_i(\hat{y}_i - \bar{y})^2}{\sum_i(y_i - \bar{y})^2}$$

$$R^2_{adj} = 1 - \left[\frac{(1 - R^2)(n - 1)}{(n - k - 1)}\right]$$

Where $y$ is the target feature, $\bar{y}$ is the mean of the observed data, $\hat{y}$ is the predicted value, $K$ is the number of independent variables and $n$ is the observations in our dataset.

In our experiments, the Carrier Origin model with the least number of trainable parameters, performed the worst. The seasonal model came second and the best performance came from the neural network which had the most trainable parameters and trained with all available data. The results can be found in Table 2.

**Table 2:** Experiment results

| Model | $R^2_{adj}$ |
|---|---|
| Model 1 – Carrier Origin | 62.45% |
| Model 2 – Seasonal | 69.3% |
| Model 3 – Neural Net | 78.64% |

**6. Limitations and Future Work**

Our solution has a number of limitations. Below, we describe some of these and suggest directions for future work. Our best model (neural net) achieved a 78.64% adjusted $R^2$ due to the challenging nature of flight delay prediction, however the results should be improved before finding use in real-world software products. Furthermore, additional testing and evaluation is needed for flights in countries other than the US as our models are trained only on the Bureau of Transportation Statistics in the US Department of Transportation dataset. Future research studies may encompass bigger datasets with datapoints from international flights and sophisticated, inventive approaches using hybrid learning and deep learning techniques that have been tuned to achieve higher performance.

**7. Conclusion**

As the national economy grows and air traffic volume rises, the flight delay problem has grown more severe in air transportation due to airspace, bases, and staff support constraints. Determining flight delays is essential since these not only cause physical discomfort for passengers and cost airlines money, but they can also lead to risky group activities. Currently, most algorithms used to anticipate flight delays are based on short samples of data, do not employ significant quantities of data in their development, and do not account for the impact of weather on delays. In this research study, three different prediction models are developed based on machine learning and deep learning; The overall models are trained on the database of the Bureau of Transportation Statistics in the US Department of Transportation. The proposed models take different inputs from the user through Alexa and then predict flight delays. The overall structure is implemented in Amazon AWS, and the backend user can obtain all the necessary information through the Kibana dashboard.

**References**

1. H. Khaksar and S. Abdolrreza. "Airline delay prediction by machine learning algorithms." *Scientia Iranica* 26.5 (2019): 2689-2702.
2. Y. J. Kim, S. Choi, S. Briceno and D. Mavris, "A deep learning approach to flight delay prediction," *2016 IEEE/AIAA 35th Digital Avionics Systems Conference (DASC)*, 2016, pp. 1-6, doi: 10.1109/DASC.2016.7778092.
3. Y. Sun, H. Song, A. J. Jara and R. Bie, "Internet of Things and Big Data Analytics for Smart and Connected Communities," in *IEEE Access*, vol. 4, pp. 766-773, 2016, doi: 10.1109/ACCESS.2016.2529723.
4. L. Belcastro, F. Marozzo, D. Talia, & P. Trunfio, "Using scalable data mining for predicting flight delays." *ACM Transactions on Intelligent Systems and Technology (TIST)* 8.1 (2016): 1-20.
5. Y. Lv, Y. Duan, W. Kang, Z. Li and F. Wang, "Traffic Flow Prediction with Big Data: A Deep Learning Approach," in *IEEE Transactions on Intelligent Transportation Systems*, vol. 16, no. 2, pp. 865-873, April 2015, doi: 10.1109/TITS.2014.2345663.
6. Z. Shu, "Analysis of Flight Delay and Cancellation Prediction Based on Machine Learning Models," *2021 3rd International Conference on Machine Learning, Big Data and Business Intelligence (MLBDBI)*, 2021, pp. 260-267, doi: 10.1109/MLBDBI54094.2021.00056.
7. R. Caruana, and N. Alexandru. "An empirical comparison of supervised learning algorithms." *Proceedings of the 23rd international conference on Machine learning*. 2006.
8. J. Schmidhuber. "Deep learning in neural networks: An overview." *Neural networks* 61 (2015): 85-117.
9. B. Divya, EP. Coyotl, and S. Gholami. "Knock, knock. Who's there? Identifying football player jersey numbers with synthetic data." *arXiv preprint arXiv:2203.00734* (2022).
10. S. Gholami, D. Byrd, FC Rodriguez, M. Kim, Y. Nakayama, M. Noori, N. Rauschmayr. "Create, train, and deploy a billion-parameter language model on terabytes of data with TensorFlow and Amazon SageMaker." *AWS Machine Learning Blog. Amazon Web Services*. 2022.
11. G. Rele, S. Gholami. "Process and add additional file formats to your Amazon Kendra Index." *AWS Machine Learning Blog. Amazon Web Services*. 2021.
12. S. Gholami. and M. Noori. "You Don't Need Labeled Data for Open-Book Question Answering." *Applied Sciences* 12.1 (2021): 111.
13. R. Brand, S. Gholami, D. Horowitz, L. Zhou, and S. Bhabesh. "Text classification for online conversations with machine learning on AWS." *AWS Machine Learning Blog. Amazon Web Services*. 2022.
14. S. Gholami, and M. Noori. "Zero-Shot Open-Book Question Answering." *arXiv preprint arXiv:2111.11520* (2021).
15. R. Balamurugan, A. V. Maria, G. Baranidaran, L. MaryGladence and S. Revathy, "Error Calculation for Prediction of Flight Delays using Machine Learning Classifiers," *2022 6th International Conference on Trends in Electronics and Informatics (ICOEI)*, 2022, pp. 1219-1225, doi:

10.1109/ICOEI53556.2022.9776709.

16. G. Gui, F. Liu, J. Sun, J. Yang, Z. Zhou and D. Zhao, "Flight Delay Prediction Based on Aviation Big Data and Machine Learning," in *IEEE Transactions on Vehicular Technology*, vol. 69, no. 1, pp. 140-150, Jan. 2020, doi: 10.1109/TVT.2019.2954094.

17. B. Zhang and D. Ma, "Flight Delay Prediciton at an Airport Using Machine Learning," *2020 5th International Conference on Electromechanical Control Technology and Transportation (ICECTT)*, 2020, pp. 557-560, doi: 10.1109/ICECTT50890.2020.00128.

18. Y. Jiang, Y. Liu, D. Liu and H. Song, "Applying Machine Learning to Aviation Big Data for Flight Delay Prediction," *2020 IEEE Intl Conf on Dependable, Autonomic and Secure Computing, Intl Conf on Pervasive Intelligence and Computing, Intl Conf on Cloud and Big Data Computing, Intl Conf on Cyber Science and Technology Congress (DASC/PiCom/CBDCom/CyberSciTech)*, 2020, pp. 665-672, doi: 10.1109/DASC-PICom-CBDCom-CyberSciTech49142.2020.00114.

19. S. Gholami, and M. Noori. "You Don't Need Labeled Data for Open-Book Question Answering." *Applied Sciences* 12.1 (2021): 111.

20. N. L. Kalyani, G. Jeshmitha, B. S. Sai U., M. Samanvitha, J. Mahesh and B. V. Kiranmayee, "Machine Learning Model - based Prediction of Flight Delay," *2020 Fourth International Conference on I-SMAC (IoT in Social, Mobile, Analytics and Cloud) (I-SMAC)*, 2020, pp. 577-581, doi: 10.1109/I-SMAC49090.2020.9243339.

21. C. Y. Yiu, K. K. H. Ng, K. C. Kwok, W. Tung Lee and H. T. Mo, "Flight delay predictions and the study of its causal factors using machine learning algorithms," *2021 IEEE 3rd International Conference on Civil Aviation Safety and Information Technology (ICCASIT)*, 2021, pp. 179-183, doi: 10.1109/ICCASIT53235.2021.9633571

22. Y. J. Kim, S. Choi, S. Briceno and D. Mavris, "A deep learning approach to flight delay prediction," *2016 IEEE/AIAA 35th Digital Avionics Systems Conference (DASC)*, 2016, pp. 1-6, doi: 10.1109/DASC.2016.7778092.

23. V. Venkatesh, A. Arya, P. Agarwal, S. Lakshmi and S. Balana, "Iterative machine and deep learning approach for aviation delay prediction," *2017 4th IEEE Uttar Pradesh Section International Conference on Electrical, Computer and Electronics (UPCON)*, 2017, pp. 562-567, doi: 10.1109/UPCON.2017.8251111.